\pdfoutput=1
\documentclass[11pt]{article}
\usepackage{acl}

\usepackage{times}
\usepackage{latexsym}

\usepackage[T1]{fontenc}
\usepackage[utf8]{inputenc}

\usepackage{microtype}
\usepackage{tipa}
\usepackage{tabularx}
\usepackage{bm}
\usepackage{mathrsfs}
\usepackage{multirow}
\usepackage{booktabs}
\usepackage{tikz}
\usepackage{tikz-qtree}
\usepackage{color}
\usepackage{xcolor}
\usepackage{makecell}
\usepackage[switch]{lineno}
\usepackage{enumitem}
\usepackage{CJKutf8}
\usepackage{bbding}
\usepackage{float}
\usepackage{pgfplots}
\usepackage{lipsum}

\title{SeSQL: Yet Another Large-scale Session-level Chinese Text-to-SQL Dataset }

\author{Saihao Huang$^1$, Lijie Wang$^{2}$, Zhenghua Li$^{1}$, Zeyang Liu$^1$, {\bf Chenhui Dou$^1$},\\ {\bf Fukang Yan$^1$}, {\bf Xinyan Xiao$^2$}, {\bf Hua Wu$^2$}, {\bf Min Zhang$^1$} \\
$^1$Institute of Artificial Intelligence, School of Computer Science and Technology, \\
Soochow University, China; $^2$Baidu Inc, Beijing, China\\
\texttt{$^1$\{shhuang21,zyliu20,chdou21,fkyan21\}@stu.suda.edu.cn}
\\\texttt{$^1$\{zhli13,minzhang\}@suda.edu.cn}
\\\texttt{$^2$\{wanglijie,xiaoxinyan,wu\_hua\}@baidu.com}}
\pgfplotsset{compat=1.17}
\begin{document}
\begin{CJK}{UTF8}{gkai}
\maketitle

\renewcommand\arraystretch{1.35}
\begin{abstract}

As the first session-level Chinese dataset, CHASE contains two separate parts, i.e., 2,003 sessions manually constructed from scratch (CHASE-C), and 3,456 sessions translated from English SParC (CHASE-T). 
We find the two parts are highly discrepant and incompatible as training and evaluation data. 
In this work, we present SeSQL, yet another large-scale session-level text-to-SQL dataset in Chinese, consisting of 5,028 sessions all manually constructed from scratch.  
In order to guarantee data quality, 
we adopt an iterative annotation workflow to facilitate intense and in-time review of previous-round natural language (NL) questions and SQL queries. 
Moreover, by completing all context-dependent NL questions, we obtain 27,012 context-independent question/SQL pairs, allowing SeSQL to be used as the largest dataset for single-round multi-DB text-to-SQL parsing. 
We conduct benchmark session-level text-to-SQL parsing experiments on SeSQL by employing three competitive session-level parsers, and present detailed analysis.

\end{abstract}
\section{Introduction}\label{sec:intro}
Text-to-SQL parsing aims to automatically transform natural language (NL) questions into SQL queries based on given databases (DBs) \cite{tang2001using}. As a key technology in an NL interface for relational DBs, it has attracted increasing attention from both academic and industrial community. Researchers have done many solid and interesting fundamental works on both dataset construction  \cite{zhong2017seq2sql,yu2018spider} and parsing model innovation \cite{zhang2019editing,wang2020rat}.

\begin{figure}[tb]
\centering
\includegraphics[width=0.48\textwidth]{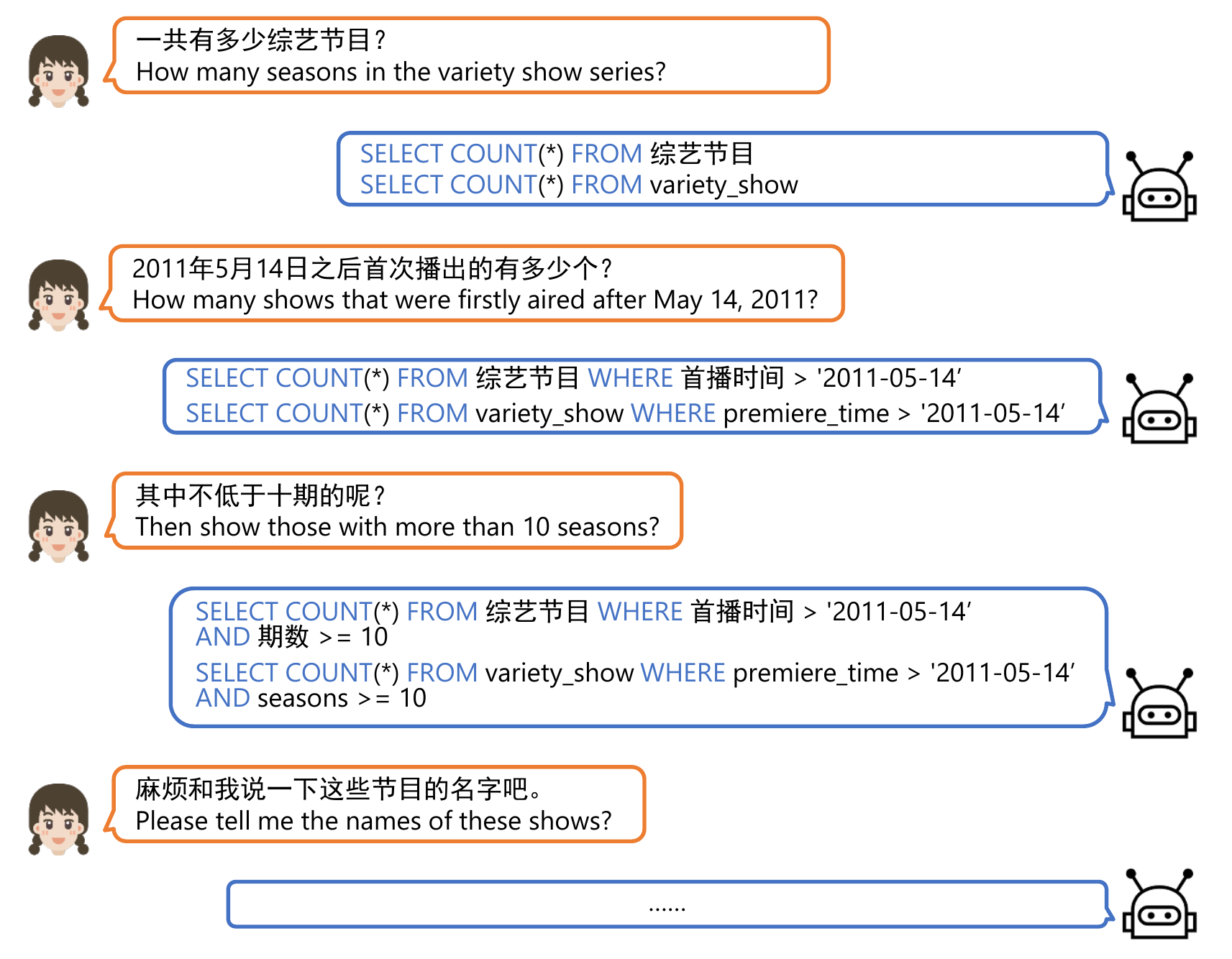}
\caption{An example session from SeSQL. 
}
\label{fig:case_intro}
\end{figure}

Previous studies mainly focus on the single-round text-to-SQL parsing, where the input questions are context-independent. 
Popular single-round datasets include WikiSQL \cite{zhong2017seq2sql} and Spider \cite{yu2018spider} for English, and 
DuSQL \cite{wang-etal-2020-dusql} for Chinese.

However, in a real-world setting, it is usually difficult for users to meet their information need via a single stand-alone question.
On the one hand, users usually have several related questions to ask at the same time, instead of a single one. 
On the other hand, possibly due to unfamiliarity toward the database or the system, users may need several trials until they find the suitable NL question.

Therefore, recent works go beyond single-round text-to-SQL parsing and start to tackle session-level text-to-SQL parsing \cite{yu2019sparc,cai2020igsql}, similar to the trend from single-round question answering (QA) to context-dependent QA  \cite{bertomeu2006contextual}. 
Figure \ref{fig:case_intro} shows a session-level example. Given a relational DB $D$, a user asks a sequence of questions, denoted by $Q = q_1, ..., q_{n}$, and the text-to-SQL engine produces a sequence of SQL queries, 
denoted by $Y = y_1, ..., y_{n}$. 
Questions in the same session are usually thematically related, and contextually dependent via ellipsis or co-reference as well \cite{bertomeu2006contextual}. 
When generating $y_j$, the parser needs to not only look at $q_j$, but also heavily rely on the previous questions.

So far, previous researchers have constructed two session-level text-to-SQL datasets, i.e., 
SParC \cite{yu2019sparc} in English and CHASE \cite{guo2021chase} in Chinese. 
SParC, containing 4,298 sessions and 12,726 question/SQL pairs, is built by extending the single-round Spider \cite{yu2018spider}.

As the first session-level Chinese dataset, CHASE contains 5,459 sessions and 17,940 question/SQL pairs \cite{guo2021chase}. 
The major problem of CHASE is that it adopted a hybrid construction method. 
Only 2,003 sessions are manually constructed from scratch (CHASE-C), whereas 3,456 correspond to a part of SParC after translating DBs and questions (CHASE-T).  
As shown in our experiments, CHASE-C and CHASE-T are highly discrepant and incompatible as training and evaluation data, possibly due to culture and language gaps. 
Moreover, only using CHASE-C may be insufficient to support model training.

This work presents \emph{SeSQL} (/\textprimstress seskju:l/), yet another large-scale session-level Chinese text-to-SQL dataset. SeSQL contains 5,028 sessions and 27,012 question/SQL pairs. All sessions are  constructed manually from scratch. 
This paper describes the construction methodology and process of SeSQL and presents detailed data analysis. 
We summarize contributions of this work as follows. 

\begin{enumerate}[label =(\arabic*),leftmargin=*]

\item SeSQL has three important features. First, based on several annotation trials, we adopt an iterative annotation workflow to encourage careful review of previous submissions, which we find is very useful for improving data quality. 
Second, we design \emph{seven categories of thematic transition} for explicitly guiding annotators to creating next-round SQL queries. Third, we follow CHASE and explicitly annotate the \emph{context-dependent types} of adjacent NL questions, such as ellipses and co-reference.

\item We complete \textcolor{black}{17,704} context-dependent questions into corresponding context-independent ones, resulting in 27,012 context-independent questions. This leads to two advantages. On the one hand, SeSQL provides the largest %Chinese 
dataset for single-round multi-DB text-to-SQL parsing. 
On the other hand, SeSQL can also support research on question completion techniques.

\item {\color{black}We conduct benchmark session-level experiments on SeSQL,} employing three competitive text-to-SQL models, i.e., EditSQL \cite{zhang2019editing}, IGSQL \cite{cai2020igsql}, and EX-RATSQL \cite{guo2021chase}.
\end{enumerate}
We will release SeSQL and the code for research  usage at \emph{{http://xyz}}.  

\section{Related Works}
\label{sec:relatedwork}
\textbf{Session-level text-to-SQL datasets.}  
To date, there exist two representative session-level text-to-SQL datasets, i.e., English 
SParC \cite{yu2019sparc} and Chinese CHASE \cite{guo2021chase}. 
SParC reuses questions in the single-round dataset Spider \cite{yu2018spider} as guidance for annotators to create question sequences. 
The basic idea is to transform an original Spider question into a sequence of simpler questions, with the goal of answering the original question. 
As pointed out by \citet{guo2021chase}, this construction method leads to two biases: 1) high proportion of context-independent questions, and 2) high proportion of easy SQL queries. 

As the first session-level Chinese dataset, CHASE is composed of two separate parts, i.e., CHASE-C and CHASE-T \cite{guo2021chase}.  
For CHASE-C, they reuse 120 DBs from single-round DuSQL\cite{wang-etal-2020-dusql}, and question/SQL pairs are created from scratch by 12 college students. 
For CHASE-T, they reuse a part of English SParC and employ 11 college students to translate DBs and question sequences into Chinese. 
However, as shown in our experiments, CHASE-C and CHASE-T exhibit different characteristics due to culture and language gaps. 
Moreover, it is inevitable that CHASE-T 
inherits the biases of SParC. 

\textbf{Conversational text-to-SQL parsing} belongs to a different task from session-level text-to-SQL parsing, and is also known as DB-based conversational QA. 
CoSQL \cite{yu2019cosql} is an English dataset for this task. 
Besides generating SQL queries, the model can ask NL questions to users  
for clarifying ambiguities.  

\textbf{Session-level text-to-SQL parsing approaches.}  
Due to space limitation, we briefly introduce four representative approaches for session-level text-to-SQL parsing. 
EditSQL \cite{zhang2019editing} generates a current-round SQL query by editing a previous-round query. Its encoder is designed to model interaction between the current-round question and all previous questions. 
IGSQL \cite{cai2020igsql} extends EditSQL by introducing a graph encoder to model DB items together with those mentioned in questions. 
\citet{hui2021dynamic} propose to jointly model the question sequence, DB items, and their interactions via a dynamic graph. 
\citet{guo2021chase} propose extended RATSQL (EX-RATSQL), a session-level variant of RATSQL \cite{wang2020rat},   
by simply concatenating all previous questions as inputs. 

\begin{figure*}[tb]
\centering
\includegraphics[width=1.01\textwidth]{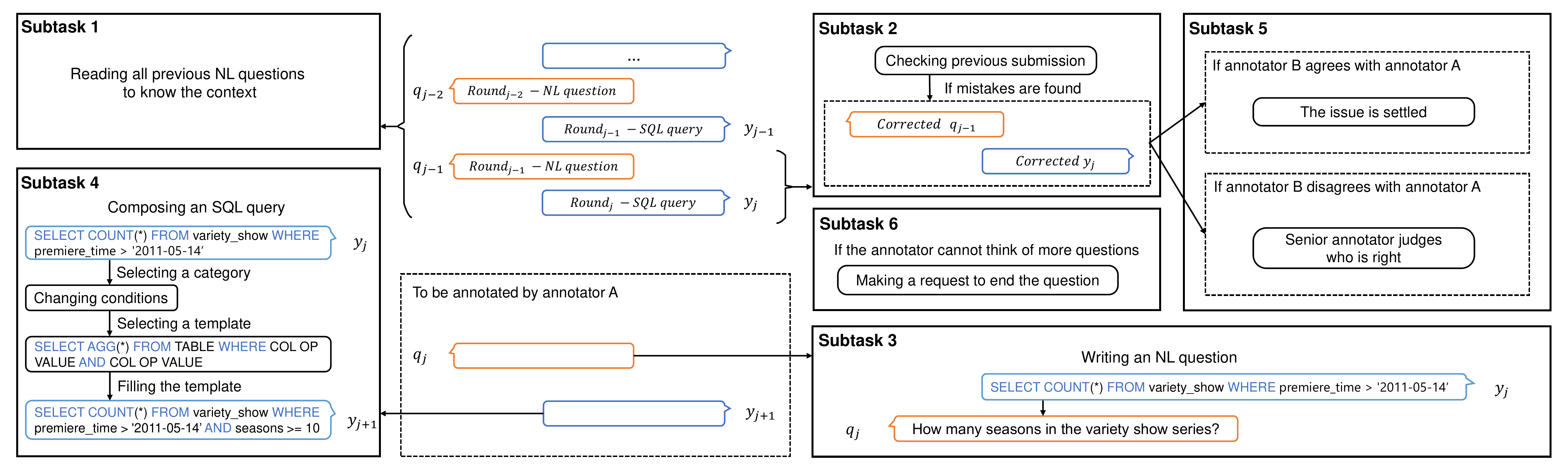}
\caption{Illustration of how to complete a current-round NL question and a next-round SQL query. }
\label{fig:interaction-system-creation-process}
\end{figure*}

\section{Dataset Construction}
\label{sec:data-construct}

The construction of SeSQL mainly consists of five steps: 1) DB  collection and cleansing, 2) initial SQL query creation, 3)  subsequent question/SQL generation, 4) review and final question creation, and 5) completing context-dependent questions. 

We first introduce our overall annotation workflow in Section  \ref{subsec:iterative-workflow}, and then detail the five steps in Sections \ref{subsec:db-cleansing}-\ref{subsec:full-review-final-question}, and finally discuss other annotation details in Section \ref{subsec:other-details}.

\subsection{An Iterative Annotation Workflow}
\label{subsec:iterative-workflow}

In the early stage of this work, we observed that one annotator tended to have a very limited number of ways for advancing a session, probably due to thinking habits and background knowledge. 
In other words, annotators usually followed a few fixed patterns to ask new questions in order to improve annotation speed. 
Therefore, if we let one annotator to complete a whole session,  the constructed data would probably contain strong  annotator-related biases \cite{Mor2019are} and be less diverse.

To deal with this issue, we adopted an \emph{iterative annotation  workflow}, as illustrated in Figure  \ref{fig:interaction-system-creation-process}. 
The basic idea is that one annotator only completes one NL  question and one SQL query, and previous submissions are  intensively reviewed by subsequent annotators. 
There are six possible subtasks for an annotator to complete at a time. 

\textbf{Subtask 1: knowing the context.} 
The annotator first reads all previous NL questions in order to 1) know what the session is about, and 2) avoid asking identical or similar questions. 

\textbf{Subtask 2: checking the previous submission.} The annotator must carefully check and correct the submission of the previous annotator, which usually consists of two parts, i.e., a current-round SQL query, and a previous-round NL question (if not first-round). 
We find that this step is very important for avoiding error accumulation. 

\textbf{Subtask 3: writing an NL question.}
The annotator writes a qualified NL question for the current-round SQL query. 
On the one hand, the question should correctly and exactly express the meaning of the SQL query. 
On the other hand, the question should be expressed  in a flexible and natural manner, imitating human conversation in real life. 

\begin{table*}[tb]
\renewcommand\arraystretch{1.25}
\renewcommand\tabcolsep{2.5pt}
\small
\centering
\scalebox{1.12} {
\begin{tabular}{l | c | c}
\toprule
\multirow{2}{*}{\textbf{Categories}} & \multicolumn{2}{c}{\textbf{Examples}} \\
\cline{2-3}
 & \makecell[c]{\textbf{current-round SQL query $y_{j}$}} & \makecell[c]{\textbf{next-round SQL query $y_{j+1}$}} \\

 \hline
Changing SELECT & \makecell[l]{\scriptsize SELECT name FROM movie} & \makecell[l]{\scriptsize SELECT name, \textbf{score}, \textbf{type} FROM movie} \\
\hline
Changing conditions & \makecell[l]{\scriptsize SELECT count(*) FROM cinema \\ \scriptsize \quad WHERE score > 3.5} & \makecell[l]{\scriptsize SELECT count(*) FROM cinema \\ \scriptsize \quad WHERE score > 3.5 AND \textbf{room\_number > 10}}\\
\hline
Changing tables & \makecell[l]{\scriptsize SELECT name FROM movie} & \makecell[l]{\scriptsize SELECT name FROM \textbf{cinema}} \\
\hline
Changing display & \makecell[l]{\scriptsize SELECT name FROM movie} & \makecell[l]{\scriptsize SELECT name FROM movie \textbf{ORDER BY score DESC LIMIT 3}} \\
\hline
Combining queries & \makecell[l]{\scriptsize SELECT name FROM cinema \\ \scriptsize \quad ORDER BY score DESC LIMIT 100} & \makecell[l]{\{\scriptsize SELECT name FROM cinema ORDER BY score DESC LIMIT 100 \} \\ \scriptsize \quad \textbf{EXCEPT} \{SELECT name FROM cinema WHERE type = "thriller"\}} \\
\hline
Hybrid of the above & \makecell[l]{\scriptsize SELECT name FROM movie} & \makecell[l]{\scriptsize SELECT name, \textbf{score} FROM movie \textbf{WHERE review\_number > 1000}} \\
\hline
Unrelated & \makecell[l]{\scriptsize SELECT name, address FROM cinema \\ \scriptsize \quad ORDER BY score DESC LIMIT 1} & \makecell[l]{\scriptsize SELECT \textbf{name} FROM \textbf{movie}} \\
\bottomrule
\end{tabular}
}
\caption{Seven categories of thematic transition 
for creating the next-round SQL query.} 
\label{tab:thematic-transition}
\end{table*}

\textbf{Subtask 4: composing an SQL query.} 
The annotator composes a new next-round SQL query, which is detailed in Section \ref{subsec:question-query-create}. 

\textbf{Subtask 5: verifying corrections via interaction. } 
If the annotator (A) finds and corrects mistakes in the previous submission of another annotator (B).  
Our annotation tool will deliver the original submission along with the corrections to annotator B for his confirmation. If annotator B agrees with A, then the issue is settled; otherwise, a senior annotator is called to make a final decision.

\textbf{Subtask 6: making a request for ending a session.}
An annotator may make a request for ending a session after completing subtask 2, when he fails to think of anything more to ask.  
Then a senior annotator handles the request.

Following \citet{yu2019sparc}, we require the number of question/SQL pairs in a session should range between 3 and 10. A session is automatically terminated if the number reaches 10.

\textbf{Discussion.} \emph{The one-annotator-one-session workflow}, adopted by CHASE,  means that a session is  completed by a single annotator.  

As we discussed in Section \ref{sec:intro}, it may introduce strong annotator-related bias, since annotator usually have a limited number of ways to advance a session. 
We observe that our iterative workflow can effectively alleviate this issue. 
Another advantage of our iterative workflow is that a previous submission is reviewed timely, which can avoid error accumulation and improve data quality. 
In contrast, data review can only be performed only after a whole session is completed in CHASE.
Nevertheless, it would be very expensive to compare the two workflows via strict quantitative experiments, which is beyond the scope of this work.

\subsection{DB Collection and Cleansing (S1)} \label{subsec:db-cleansing}

Collecting DBs is a non-trivial work. For simplicity, we reuse all 201 DBs with 813 tables of the DuSQL dataset\footnote{The license and data is at \url{https://www.luge.ai/\#/luge/dataDetail?id=13}.}  \cite{wang-etal-2020-dusql}.

After looking into the data, we find that there are a lot of noises in the original DBs of DuSQL, which is also pointed out by \citet{guo2021chase}.
Most noises fall into four categories: 1) primary or foreign keys are not given; 2) the value type of a cell does not match its column type; 
3) some cells do not have values; 4) a duplicate value occurs in the primary key column. 

In order to improve the quality of DBs and make sure that all legal SQL queries can be successfully executed,  
our six senior annotators have manually checked and corrected all DBs\footnote{Please note that we ask annotators not to introduce identification information and ask them to anonymize the existing identification information.} before real annotation. Each annotator handles about 35 DBs.

\subsection{Initial SQL Query Creation (S2)}

Creating suitable initial queries are crucial for session-level text-to-SQL data creation, since they directly influence subsequent annotations. 
The suitability of initial queries depends on two aspects, i.e., simplicity and diversity. 
Regarding to the first aspect, we find that queries at easy and medium difficulty levels are the most appropriate as initial queries. 
We follow definition of difficulty levels in \citet{yu2018spider}. 
The second aspect indicates that initial queries should cover as many SQL keywords as possible. 
In order to satisfy both aspects, we induced 60 SQL query templates from single-round Spider and DuSQL, and each template contains some slots corresponding to masked table/column names and cell values. 
Given a DB, we require the initial query matches one of the templates (simplicity), and create at most  one initial query for one template (diversity). 

We create 5,028 valid initial queries in total, among which 1,761 are from DuSQL, and 3,267 are written from scratch by our senior annotators.

\subsection{Subsequent Question/SQL Creation (S3)} 
\label{subsec:question-query-create}

As discussed in Section \ref{subsec:iterative-workflow}, subsequent questions and queries are created by multiple randomly selected annotators, each contributing one current-round NL question and one next-round SQL query. 
This subsection focuses on how to create a next-round SQL query given existing context.  

Similar to context-dependent QA \cite{bertomeu2006contextual}, it is crucial to make as realistic as possible the thematic transition and context dependency between adjacent utterances, where theme refers to users' information need, and context dependency is concerned with manners in reusing previous content.
In this step (i.e., S3), we mainly consider the thematic transition, since reusing previous content is usually a natural choice for annotators. 
As for the context dependency information, we follow CHASE and create explicit annotation (see Section  \ref{ssec:fine-annotation}).

\textbf{Seven categories of thematic transition.}
To capture theme change and encourage diversity, we design seven transition categories to represent the relationship between the current-round and next-round queries, i.e., $y_j$ and $y_{j+1}$, as illustrated in Table \ref{tab:thematic-transition}. 
Please note that we also allow annotators to compose a thematically ``unrelated'' query, which sometimes happens in real-world scenario. 

Figure \ref{fig:interaction-system-creation-process} illustrates concrete operations in the bottom left corner. 
Given $y_j$, the annotator first selects a transition category; then our annotation tool suggests several potential SQL templates according to $y_j$ and the selected category; finally the annotator selects an SQL template and fills it with DB elements to complete an SQL query.

\subsection{{Review \& Final Question Creation (S4)}}
\label{subsec:full-review-final-question} 

\begin{table*}[tb]
\small
\renewcommand\tabcolsep{2.5pt}
\centering
\scalebox{0.93} {
\begin{tabular}{l | c c c c c c | c c | c c | c}
\toprule
\multirow{2}{*}{Datasets} & \multicolumn{6}{c|}{General Information} & \multicolumn{2}{c|}{Challenge Information} & \multicolumn{2}{c|}{Contextual Annotation} & \multirow{2}{*}{\makecell[c]{Question \\ Completion}}\\
\cline{2-7} \cline{8-9} \cline{10-11}
 & Language & DBs & Tables & Sequences & Pairs & Avg. round & Dep. Ratio & Easy Ratio & Thematic & Dependency & \\
\hline
SparC & English & 200 & 1,020 & 4,298 & 12,726 & 3.0 & 52.5 & 40.1 & \XSolidBrush & \XSolidBrush & \XSolidBrush \\
\hline  \hline
CHASE & Chinese & \textbf{280} & \textbf{1,280} & \textbf{5,459} & 17,940 & 3.3 & 64.7 & 27.7 & \Checkmark & \Checkmark & \XSolidBrush \\
\quad  CHASE-C &  & 120 & 462 & 2,003 & 7,694 & 3.8 & \textbf{71.2} & 18.6 &  &  &  \\
\quad  CHASE-T &  & 160 & 818 & 3,456 & 10,246 & 3.0 & 57.8 & 37.4 &  &  &  \\
\hline
SeSQL & Chinese & 201 & 813 & 5,028 & \textbf{27,012} & \textbf{5.4} & 65.5 & \textbf{13.6} & \Checkmark & \Checkmark & \Checkmark\\
\bottomrule
\end{tabular}
}
\caption{Statistics of existing cross-domain context-dependent datasets. ``Avg. round'' represents the average number of rounds in a sequence. ``Dep. Ratio'' represents the ratio of the context-dependent questions, and ``Easy ratio'' represents the ratio of queries in the easy level. Please note that CHASE only partially released coarse-grained thematic transition relations.} 
\label{tab:data_all_stas}
\end{table*}

This step is performed by our senior annotators. 
If an ordinary annotator makes a request to terminate a session, the annotation tool will transmit the request to a senior annotator.
If the senior annotator agrees, then he must carefully review all previous questions and SQL queries, and correct all found mistakes. 
After that, he writes an NL question for the final-round SQL query.

\subsection{Completing Context-Dep Questions (S5)
}
\label{ssec:fine-annotation}

In order to capture context dependency and make our dataset more widely applicable, we perform this step in a separate manner after all sessions are completed via the above four steps  (S1-S4).

Each session is then assigned to one senior annotator. The annotator first goes through all NL questions, and decides whether each question is context-dependent. 
Then, the context-dependent question is rewritten into a corresponding context-independent one. 
There are in total 17,704 context-dependent questions, accounting for 65.5\% of all questions (see Table \ref{tab:dep_relation}). 
As a result, \emph{SeSQL can serve as a single-round Chinese text-to-SQL dataset as well}, like DuSQL \cite{wang-etal-2020-dusql}. Moreover, it can also support reserach on question completion techniques.

\textbf{Context dependency types.} 
Inspired by CHASE \cite{guo2021chase} and context-dependent QA \cite{bertomeu2006contextual}, 
we ask annotators to explicitly annotate the way that a context-dependent question depends on its previous questions. 
There are five types, i.e., independent, co-reference, ellipsis, hybrid of co-reference and ellipsis, and others. 
Such annotation can help us to better understand results of text-to-SQL parsers.  

\subsection{Other Annotation Details}
\label{subsec:other-details}

\textbf{Annotators and Training.} 
We recruit 28 undergraduate students as our part-time annotators, and 6 master students as senior annotators, including three co-authors of this paper. All of them come from the computer science department of our university and are familiar with the SQL language.  

Before real annotation, we train all annotators for several times so that they understand the text-to-SQL parsing task, the annotation workflow, and the annotation tool, etc. During real annotation, we have also held several meetings to discuss common mistakes and settle disputes. 
Our annotation project lasts for about half a year. 

\textbf{Annotation tool.} We build an online browser-based annotation tool to facilitate this work. Figure \ref{fig:annotation_tools} in Appendix \ref{sec:annotation-tool} shows the annotation interface.

\textbf{Payment.} 
All annotators were paid for their work based on the quality and quantity of their annotations.
According to the annotation time recorded by our annotation tool, the average salary per hour is 25 RMB for ordinary annotators, and 35 RMB for senior annotators.\footnote{The average salary is about 20 RMB for a part-time KFC employee in our city.}
A total of 106K RMB is paid to annotators. 

\section{Analysis of SeSQL}
\label{sec:data-analysis}
\textbf{Basic statistics.} 
As shown in Table \ref{tab:data_all_stas}, SeSQL contains 5,028 unique question sequences over 201 DBs, with 27,012 questions annotated with their corresponding SQL queries. 
First of all, compared with English text-to-SQL datasets, both SeSQL and CHASE have a larger number of sessions and question/query pairs. 
Second, both SeSQL and CHASE are more challenging, due to higher percentage of context-dependent and non-easy questions.

Compared with CHASE, SeSQL contains more question/query rounds per session. 
We believe this owes to the seven categories of thematic transition that we design, which makes it more flexible for annotators to create next-round SQL queries. 
Moreover, SeSQL has overall higher percentages of context-dependent and non-easy questions. 

Looking into CHASE, as we earlier discussed, only 2,003 sessions and 7,694 question/query pairs (i.e., CHASE-C) are annotated from scratch, which are much fewer than SeSQL. In the experiments, we show that CHASE-C and CHASE-T are highly discrepant and incompatible as  training and evaluation data. 

Finally, SeSQL provides corresponding context-independent questions for all context-dependent ones, and thus can also serve as a single-round text-to-SQL dataset.

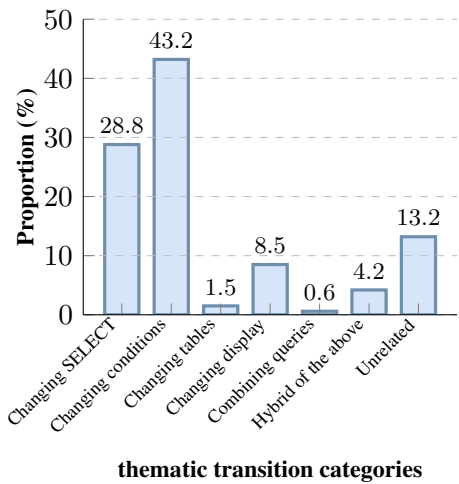
\begin{figure}[tb]
  \centering
  \begin{tikzpicture}
    \begin{axis} [width=6.5cm, height=5.5cm,
      ybar,
      bar width=0.45cm,
      axis on top,
      axis x line*=bottom,
      axis y line*=left,
      enlarge x limits=0.13,
      tick align=inside,
      ymin=0, ymax=50,
      ylabel={Proportion (\%)},
      ylabel style = {yshift=-0.5em,font=\small\bfseries},
      ytick={0, 10, ..., 50},
      symbolic x coords={
        Changing SELECT, Changing conditions, Changing tables, Changing display, Combining queries, Hybrid of the above, Unrelated,},
      xtick=data,
      nodes near coords,
      nodes near coords align={vertical},
      nodes near coords style={
      /pgf/number format/.cd,
      fixed,
      fixed zerofill,
      precision=1,
      /tikz/.cd
      },
      xlabel={thematic transition categories},
      xlabel style = {yshift=-0.0em,font=\small\bfseries},
      ymajorgrids=true,
      grid style=dashed,
      x tick label style={rotate=45,anchor=east,font=\scriptsize},
    ]
    \addplot [draw={rgb,255:red,76; green,114; blue,150},
    %\addplot [draw={rgb,255:red,215; green,233; blue,247},
    very thick,
    draw opacity=0.8,
    font=\small\bfseries,
     fill={rgb,255:red,214; green,229; blue,250}] coordinates {
      (Changing SELECT, 28.8)
      (Changing conditions, 43.2)
      (Changing tables, 1.5)
      (Changing display, 8.5)
      (Combining queries,0.6)
      (Hybrid of the above, 4.2)
      (Unrelated, 13.2)
    };
\end{axis}
\end{tikzpicture}
\caption{Thematic transition distribution in SeSQL.}
\label{fig:thematic_dis}
\end{figure}

\textbf{Thematic transition.}
We compute the thematic transition distributions in Figure \ref{fig:thematic_dis}. We find that the most frequently occurring transitions are \emph{Changing conditions} and \emph{Changing SELECT}, which are two very common contextual thematic relations in conversational QA systems \cite{bertomeu2006contextual}. Meanwhile, it can be seen that transitions of \emph{Changing tables} and  \emph{Combining queries} rarely occur. In the former case, the next-round SQL query raises another related topic, and its NL question is usually context-independent. The latter case usually leads to a very complex next-round SQL query. 
\begin{table}[tb]
\small
\center
\scalebox{0.95} {
\begin{tabular}{l | c c c c c}
\toprule
Datasets & Indep. & Core. & Elli. & Both & Others\\
\hline
SParC & \textbf{47.5} & 31.6 & 25.9 & 5.0 & 0 \\
CHASE & 35.3 & 35.7 & 28.5 & 0.5 & 0 \\
\quad CHASE-C & 28.8 & \textbf{39.8} & 30.9 & 0.5 & 0 \\
\quad CHASE-T & 42.2 & 31.4 & 24.7 & 1.7 & 0 \\
SeSQL & 34.5 & 13.4 & \textbf{35.0} & \textbf{12.5} & \textbf{4.6} \\
\bottomrule
\end{tabular}
}
\caption{Context dependency distributions of existing datasets, where the reported results of SParC, CHASE, CHASE-C and CHASE-T are from \citet{guo2021chase} .}
\label{tab:dep_relation}
\end{table}

\textbf{Context dependency.} Table \ref{tab:dep_relation} shows  distribution of context dependency types of different datasets. Following previous studies, there are five types of context dependency, i.e.,  independent (indep.), co-reference (core.), ellipsis (elli.), hybrid of co-reference and ellipsis (both) \footnote{The values of ``both'' for other three datasets are inferred from their reported results of ``Core.'' and ``Elli.''.}, and others. 
First, the proportion of context-independent questions in SeSQL is much lower than SParc and CHASE-T, and is only 5.7\% higher than CHASE-C.  
Second, SeSQL has the highest percentage of questions with ellipsis, and of questions with both co-reference and ellipsis. 
Finally, the remaining 4.6\% of questions in SeSQL are related with previous questions in other ways than co-reference and ellipsis.  
From the distribution analysis, we believe that compared with CHASE,  SeSQL can be used as a new and complementary resource for research on session-level text-to-SQL parsing.  

\begin{table}[tb]
\small
\center
\begin{tabularx}{0.36\textwidth}{l | r r r}
\toprule
Split Set & \# DB & \# Sequence & \# Pair \\
\hline
Train & 160 & 4,002 & 21,454 \\
Dev & 17 & 425 & 2,279 \\
Test & 24 & 601 & 3,279 \\
\bottomrule
\end{tabularx}
%}
\caption{Dataset split statistics.}
\label{tab:data-split}
\end{table}

\section{Experiments}
\label{sec:experiment}
\textbf{Datasets.}
According to the cross-domain setting, we split SeSQL such that there is no DB overlap in train/dev/test sets. Since our DBs are from DuSQL, we follow its DB split for three sets of SeSQL. Table \ref{tab:data-split} shows the data split statistics.

\textbf{Evaluation metrics.}
We use two popular metrics to evaluate model performances: Question-level Match (QM), the exact matching score over all questions, and Interaction-level Match (IM), the exact matching score over all interactions.
The exact matching score is 1 for a question only if all its predicted SQL clauses are correct, and 1 for an interaction only if the exact matching score for every question in the interaction is 1.

\textbf{Benchmark approaches.}
We adopt several competitive models that have published the corresponding source codes as the baseline approaches, i.e., EditSQL \cite{zhang2019editing}, IGSQL \cite{cai2020igsql} and extended RATSQL (EX-RATSQL) \cite{guo2021chase} for the context-dependent setting, as well as RATSQL \cite{wang2020rat} and LGESQL \cite{cao2021LGESQL} for the context-independent setting.
%where EX-RATSQL and LGESQL have achieved SOTA results on English datasets. 
Due to space limitation, we show their implementation details in Appendix \ref{sec:base-model}.

\begin{table}[tb]
\small
\center
\scalebox{0.88} {
\begin{tabular}{l |c c | c c}
\toprule
\multirow{2}{*}{Models} & \multicolumn{2}{c|}{QM} & \multicolumn{2}{c}{IM} \\
\cline{2-3} \cline{4-5}
 & Dev & Test & Dev & Test \\
\hline
EditSQL \cite{zhang2019editing} & 57.2 & 52.6 & 27.3 & 22.6 \\
IGSQL \cite{cai2020igsql} & 63.3 & 59.5 & \textbf{35.0} & \textbf{29.0} \\
EX-RATSQL \cite{guo2021chase} & 56.6 & 50.4 & 18.9 & 17.0 \\
\hline
RATSQL \cite{wang2020rat} & 65.6 & 56.5 & - & - \\
LGESQL \cite{cao2021LGESQL} & \textbf{76.8} & \textbf{71.0} & - & - \\
\bottomrule
\end{tabular}
}
\caption{Performances of base approaches over all questions (QM) and all interactions (IM). All reported results are the average over two runs.}
\label{tab:multi_main_result}
\end{table}

\begin{table}[tb]
\small
\center
\scalebox{0.85} {
\begin{tabular}{l | r r | r r | r r}
\toprule
\multirow{2}{*}{Training Data} & \multicolumn{2}{c|}{SeSQL} & \multicolumn{2}{c|}{CHASE-C} & \multicolumn{2}{c}{CHASE-T}\\
\cline{2-3} \cline{4-5} \cline{6-7}
 & QM & IM & QM & IM & QM & IM \\
\hline
CHASE & \textbf{12.9} & \textbf{0.3} & 33.4 & 9.9 & \textbf{43.9} & \textbf{24.6}\\
CHASE-C & 4.0 & 0.1 & \textbf{35.7} & \textbf{11.0} & 22.5 & 6.1\\
CHASE-T & 12.1 & \textbf{0.3} & 19.4 & 2.5 & 42.3 & 23.8\\
\hline
SeSQL & 61.7 & 32.9 & 22.0 & 4.0 & 30.4 & 15.0\\
SeSQL + CHASE-C & 62.5 & 33.2 & \textbf{39.3} & \textbf{14.0} & 33.8 & 15.4 \\
SeSQL + CHASE-T & \textbf{63.8} & \textbf{34.2} & 24.6 & 3.7 & \textbf{46.5} & \textbf{28.5}  \\
\bottomrule
\end{tabular}
}
\caption{Performances of IGSQL on dev sets of three datasets using different training data. All results are the average over three runs.}
\label{tab:inter-chase-ours}
\end{table}

\begin{table*}[tb]
\small
\renewcommand\tabcolsep{2.5pt}
\centering
\scalebox{0.93} {
\begin{tabular}{l | c c c c c c c | c c c c c | c c c c c}
\toprule
\multirow{2}{*}{Models} & \multicolumn{7}{c|}{Thematic Transition} & \multicolumn{5}{c|}{Context Dependency} & \multicolumn{5}{c}{Round Number} \\
\cline{2-8} \cline{9-13} \cline{14-18}
 & SEL. & Cons. & Tab. & Display & Com. & Hybrid & Unrel. & Indep. & Core. & Elli. & Both & Others & 1 & 2 & 3 & 4 & $\ge$5 \\
\hline
EditSQL & 51.3 & 47.6 & 40.7 & 50.4 & 40.0 & 31.0 & 59.6 & 63.2 & 51.2 & 45.4 & 47.3 & 51.0 & 65.4 & 57.3 & 50.7 & 45.4 & 46.9\\
IGSQL & \textbf{59.5} & \textbf{55.0} & 40.7 & \textbf{59.8} & \textbf{55.0} & \textbf{43.0} & \textbf{65.6} & 67.4 & \textbf{59.3} & \textbf{53.4} & \textbf{56.0} & \textbf{60.3} & 68.6 & \textbf{64.6} & \textbf{58.1} & \textbf{55.2} & \textbf{53.9}\\
EX-RATSQL & 48.1 & 43.5 & \textbf{51.2} & 48.2 & 45.0 & 37.0 & 58.9 & 63.0 & 47.3 & 41.9 & 44.4 & 50.6 & 64.8 & 58.0 & 46.6 & 42.4 & 44.1\\
\bottomrule
\end{tabular}
}
\caption{Fine-grained QM results on the test set of SeSQL. All reported results are the average over two runs.}
\label{tab:multi_fine_result}
\end{table*}

\subsection{Results}
\label{ssec:result}

\textbf{Overall performances.} Table  \ref{tab:multi_main_result} shows the overall performances of five baseline models, where the first row shows performances of three session-level models (i.e., EditSQL, IGSQL and EX-RATSQL) on the SeSQL's session-level data, and the second row shows performances of RATSQL and LGESQL on the single-round data of SeSQL.  
IGSQL and LGESQL have achieved the best performances on the session-level and single-round data, respectively.
But the results on the session-level data are far from satisfactory, reflected in two aspects. First, the best performances on IM, the primary metric in the session-level setting, only achieves 29.0\% on the test set. Second, the best QM accuracy achieved by IGSQL is 59.5\%, where the best QM accuracy on the single-round data is 71.0\%. That is, there is a large room for both QM and IM improvements on SeSQL. We believe SeSQL can facilitate the research on session-level text-to-SQL parsing.

\textbf{Comparison between SeSQL and CHASE.} 
We use different combination of SeSQL and CHASE as training data, and use three separate dev sets, in order to understand data similarity and discrepancy.  
To avoid DB overlap, which would corrupt the cross-DB text-to-SQL parsing task, 
we remove all DBs that also appears in any of the three dev sets from each training data, along with corresponding question/SQL pairs. 
Table \ref{tab:inter-chase-ours} shows the results.

First of all, it is clear that CHASE-C and CHASE-T are highly discrepant and incompatible. Using whole CHASE as training data leads to performances drop on CHASE-C dev set, compared with using only CHASE-C. 
In other words, the extra CHASE-T only introduces more noisy information than helpful information. 
However, using whole CHASE increases performances on CHASE-T dev set, compared with using only CHASE-T.  

Second, using only SeSQL as training data achieves acceptable cross-dataset performances on both CHASE-C dev set, which is much higher than using CHASE-T as training data. The same trend goes to CHASE-T dev set. 
This indicates that SeSQL possesses a higher level of generalization ability.

Third, using both SeSQL and CHASE-C as training data leads to higher performances on CHASE-C dev set than using only CHASE-C. 
Similarly, using both SeSQL and CHASE-T as training data leads to higher performances on CHASE-T dev set than using only CHASE-T as well. 
Such consistent improvement indicates that SeSQL is of higher quality and compatible with both CHASE-C and CHASE-T. 

Finally, using either CHASE-C or CHASE-T as extra training data increases QM and IM on SeSQL dev set slightly, compared with using only SeSQL. 
We suspect this may be due to the increased data volume added by both datasets.

\textcolor{black}{Despite SeSQL improves cross-dataset generalization, the model generalization ability across different datasets is still weak, even if these datasets are built on the same DBs (e.g., SeSQL and CHASE-C). We believe SeSQL can facilitate the research of text-to-SQL parsing, especially on the cross-dataset generalization of text-to-SQL models.}

\subsection{Analysis} 
\label{ssec:exp-analysis}
According to the fine-grained annotation information, we report QM results on SeSQL's test set in Table \ref{tab:multi_fine_result}. There are three main findings that are applicable to all baseline models.
First, among all thematic transitions, all models do not perform well on transitions of \emph{Combining queries} (Com.) and hybrid of other transitions. As described in Section \ref{sec:data-analysis}, these transitions usually result in complex query generation. 
Second, as shown in the column of ``\emph{Context Dependency}'', QM performances on context-independent (Indep.) pairs is higher than that on context-dependent pairs, i.e., the other four dependency types. \textcolor{black}{Furthermore, all base models do not perform well on questions that omit important historical information, i.e., labeled as ``ellipse''. This proves that how to effectively use historical information is challenging}.
Finally, due to the difficulty increase in SQL generation, QM performances decreases as the round increases, \textcolor{black}{which is consistent with the conclusions in other session-level datasets, e.g., SParC and CHASE}. 

Then we analyze significance of fine-grained annotations
by comparing different models. From Table \ref{tab:multi_fine_result}, there are three interesting findings to verify the importance of fine-grained annotations in revealing the effectiveness of model components.  
First, both IGSQL and EditSQL perform better than EX-RATSQL on non-first round questions, as they refer to the previous-round SQL query during the generation of the current-round SQL query. As all we known, in the context-dependent setting, the historical questions and generated SQL queries are very important to the current SQL generation.
Second, IGSQL outperforms EditSQL on all transition types and dependency types, where IGSQL incorporates a graph encoder into EditSQL to model DB schema items together with items mentioned in historical questions. The performances on these fine-grained annotations verify that this graph encoder effectively captures historical information of questions and DB schema items.
Third, EX-RATSQL, which only uses a relation-aware transformer to model the historical questions and DB schema items, performs best on the transition type of \emph{Tab.}, in which there is weak correlation between the historical rounds and the current round.
Based on the above conclusions, we believe these annotations can help to reveal the advantages and limitations of the model, so as to help to improve models.
\section{Conclusions}
This paper presents SeSQL, yet another large-scale session-level Chinese text-to-SQL dataset. 
We describe its construction methodology and process in detail, and present detailed analysis about it. 
We conduct benchmark experiments with three representative session-level parsers, and prove that SeSQL exhibits several important features compared with CHASE.
First, all 5,028 sessions are manually constructed from scratch, whereas only 2,003 sessions in CHASE-C are manually constructed from scratch. 
Second, being used as extra training data, SeSQL can consistently improve performances on both CHASE-C and CHASE-T. This indicates SeSQL is of higher quality and has stronger generalization ability. 
Third, by completing context-dependent questions, SeSQL provides 27,012 context-independent question/SQL pairs, and thus can be used as a solid dataset for future research on single-round text-to-SQL parsing. 
%\input{content/acknowledgements.tex}

\bibliography{arxiv}
\bibliographystyle{acl_natbib}
\clearpage
\appendix

\appendix

\vspace{+1ex}
\begin{center}
\Large \textbf{Appendices} 
\end{center}

\vspace{+1ex}

\begin{figure*}[tb]
\centering
\includegraphics[width=1\textwidth]{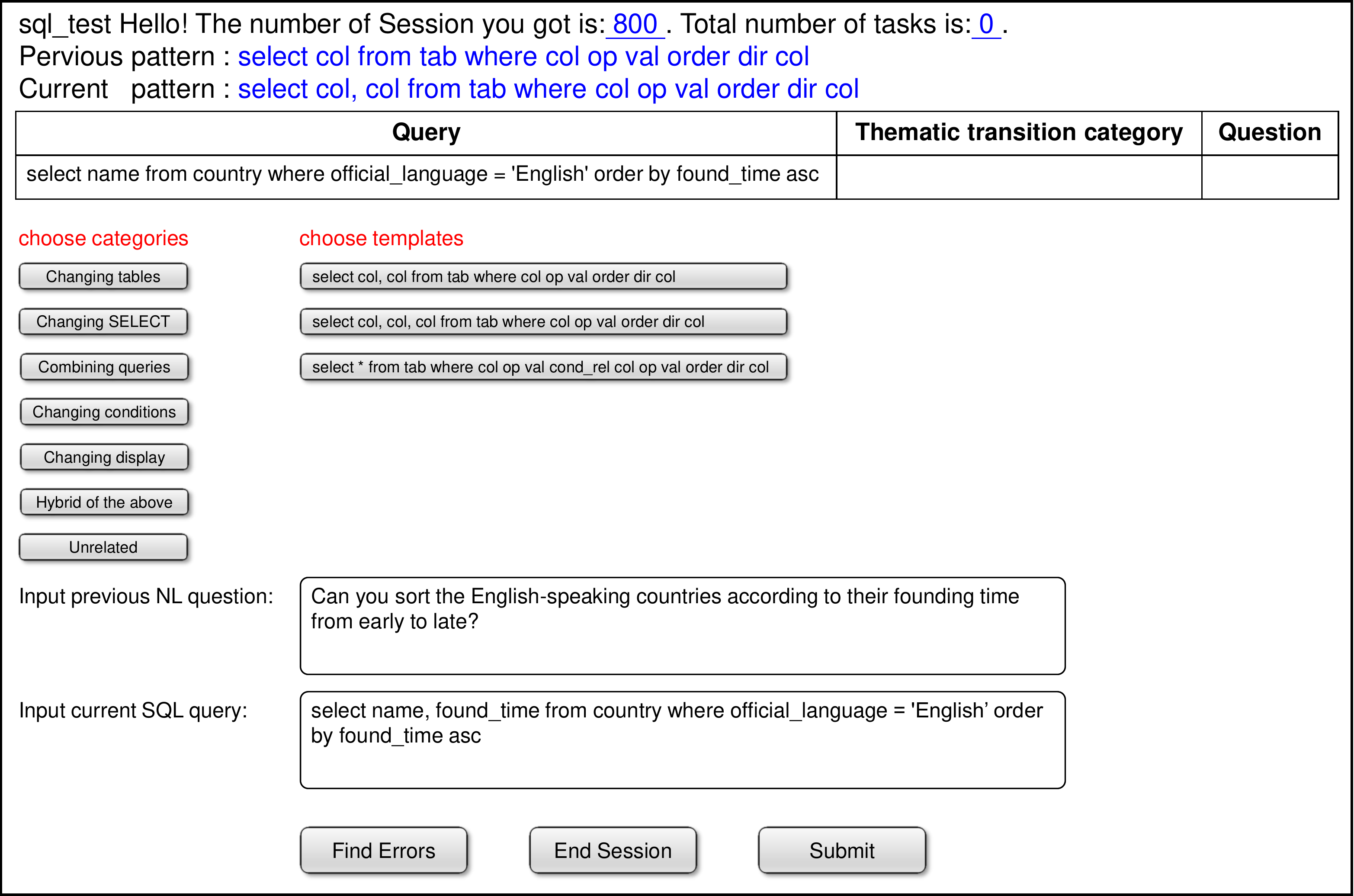}
\caption{User interface of our annotation tool.}
\label{fig:annotation_tools}
\end{figure*}

\section{Benchmark Approaches}
\label{sec:base-model}

In the context-dependent setting, many researchers focus on parsing model innovation \cite{yu2020score, hui2021dynamic, zheng2022hie}. In this work, we adopt Edit-SQL \cite{zhang2019editing}, IGSQL \cite{cai2020igsql} and extended RATSQL (EX-RATSQL) \cite{guo2021chase} as our benchmark approaches to evaluate their performances on our new dataset - SeSQL. For the three baseline approaches, we use default parameter settings in their released codes. 
Please note that in our experiments, we use fixed values given in the released source codes for hyper-parameters, that is, we do not perform hyper-parameter search to find best hyper-parameter values.
Meanwhile, following \citet{guo2021chase}, we use BERT-base \cite{devlin2019bert} to enhance the three parsers.

\textbf{Edit-SQL}\footnote{\url{https://github.com/ryanzhumich/editsql}} utilizes the interaction history by editing the previous predicted query to improve the generation quality, as the adjacent NL questions are often linguistically dependent and their corresponding SQL queries tend to overlap. In the decoding process, they view an SQL query as a sequence and use an editing mechanism to reuse the previous generated SQL query at the token level.
In the encoder, in order to deal with complex table structures, they employ an question-table encoder to incorporate the context of the user question and the table schema.

It takes about 7 days to train a basic BERT enhanced EditSQL model on a V100 GPU card. The EditSQL model has about 115M parameters.

\textbf{IGSQL}\footnote{\url{https://github.com/headacheboy/IGSQL}} incorporates a graph encoder into EditSQL to capture historical information of user questions and database schema items. 
In encoding phase, they not only use an interaction encoder to capture historical information of user NL questions, but also use a DB schema interaction graph encoder to utilize historical information of DB schema items.
In decoding phase, for making the prediction of SQL tokens, IGSQL introduces a gate mechanism to weigh the importance score of vocabularies from different sources, including DB schema items and the previous generated SQL query.

With a V100 GPU card, it takes about 6 days to train a basic BERT enhanced IGSQL model. And an IGSQL model has about 110M parameters.

\textbf{EX-RATSQL}\footnote{\url{https://github.com/xjtu-intsoft/chase/tree/main/Benchmark_Approaches/DuoratChar}} is an extension of RATSQL \cite{wang2020rat}, which uses a relation-aware transformer encoder and a gramma-based decoder to generate SQL Queries and performs well in the context-independent setting. Compared with RATSQL, EX-RATSQL adopts a simple concatenation context modeling approach to concatenate current question and all prior context-dependent questions with a special symbol [SEP] from the back forward. In other respects, it keeps up with RATSQL. The code of EX-RATSQL is based on DuoRAT \cite{scholak2021DuoRAT} and we set batch size as 12 and max steps as 200,000 in our experiments. Besides it, we use default values for other hyper-parameters.

It also takes about 8 days to train a basic BERT enhanced EX-RATSQL model on a V100 GPU card. An EX-RATSQL model has about 135M parameters.

In the context-independent setting, we use a widely-used approach RATSQL \cite{wang2020rat} and a SOTA approach LGESQL \cite{cao2021LGESQL} as the benchmark approaches to understand the characteristics of our dataset. We evaluate RATSQL and LGESQL on all context-independent pairs.

\textbf{RATSQL}\footnote{\url{https://github.com/microsoft/rat-sql}} utilizes a relation-aware transformer encoder to better model the connections between DB schemas and NL questions. Then it uses a grammar-based decoder to ensure the grammaticality of the generated SQL queries. In our experiments, we set batch size as 4 and random initialize seed as 0. We use default values for other parameter in the released code, and use the fixed values (listed in the released code) for hyper-parameters. Similarly, we use Chinese BERT-wwm \cite{cui2021pre} to enhance our RATSQL model. 

Training RATSQL is also expensive. It takes about 6 days to train a basic BERT-wwm enhanced RATSQL model on a V100 GPU card. A RATSQL model has about 168M parameters.

\textbf{LGESQL}\footnote{\url{https://github.com/rhythmcao/text2sql-lgesql.git}} takes advantages of Dual RGAT to jointly encode the questions and the schemas. Compared with RATSQL, LGESQL pays more attention to the topological structure of edges and applies an edge-centered line graph to enhance the encoding of 1-hop edge features.  In addition, it comes up with a graph pruning method as an auxiliary task to help the encoder improve the discriminative capability. Later, it uses a grammar-based decoder to generate SQL Queries as well. In our experiments, we leave the default hyperparameters unchanged.

Training a basic BERT enhanced LGESQL model spends 8 days on a V100 GPU card. The LGESQL model has about 148M parameters.

\section{Annotation Tool}
\label{sec:annotation-tool}
Figure \ref{fig:annotation_tools} shows the user interface of our annotation tool. The details of annotating process have been mentioned in Section \ref{sec:data-construct}. During the process of annotating and checking question/SQL sequences, this annotation tool helps to improve annotation speed and data quality.

\section{More Annotation Examples}
\label{sec:more-examples}

\begin{table*}[tb]
\renewcommand\tabcolsep{2.5pt}
\renewcommand\arraystretch{1.3}
\small
\centering
\scalebox{0.624} {
\begin{tabular}{l l c c}
\toprule
\textbf{\#} & \textbf{Question\&SQL Query} & \textbf{Context Dependency} & \textbf{Thematic Transition}\\
\hline
\multicolumn{2}{l}{\textbf{Session 1}}\\
\hline
\multirow{2}{*}{$q^{1}_{1}$} & {请列出所有频道的相关收视信息。} & \multirow{4}{*}{\makecell[c]{Independent}} & \multirow{4}{*}{\makecell[c]{--}}\\
~ & (Please list the relevant information of all channel.) \\ \multirow{2}{*}{$y^{1}_{1}$} & {select * from 频道收视}\\ ~ & (select * from channel\_ratings)\\
\hline
\multirow{2}{*}{$q^{1}_{2}$} & {市场份额最高的那个呢？} & \multirow{4}{*}{\makecell[c]{Dependent \\ (Ellipsis)}} & \multirow{4}{*}{\makecell[c]{Changing display}}\\  ~ & (Tell me the one with the highest market share.)\\ \multirow{2}{*}{$y^{1}_{2}$} & select * from 频道收视 order by 市场份额 desc limit 1\\ ~ & (select * from channel\_ratings order by market\_share desc limit 1) \\
\hline
\multirow{2}{*}{$q^{1}_{3}$} & {还是把市场份额不低于10\%的都列出来吧。} & \multirow{4}{*}{\makecell[c]{Dependent \\ (Ellipsis)}} & \multirow{4}{*}{\makecell[c]{Changing conditions}}\\ ~ & (Now list all relevent information about channels with a market share of no less than 10\% instead.) \\ \multirow{2}{*}{$y^{1}_{3}$} & {select * from 频道收视 where 市场份额 >= 0.1}\\ ~ & (select * from channel\_ratings where market\_share >= 0.1)\\
\hline
\multirow{2}{*}{$q^{1}_{4}$} & {其中直播关注度超过0.1\%的呢？} & \multirow{4}{*}{\makecell[c]{Dependent \\ (Both)}} & \multirow{4}{*}{\makecell[c]{Changing conditions}}\\ ~ & (How about counting only those in the above results with more than 0.1\% live streaming attention?)\\ \multirow{2}{*}{$y^{1}_{4}$} & {select * from 频道收视 where 市场份额 >= 0.1 and 直播关注度 > 0.001}\\~ & (select * from channel\_ratings where market\_share >= 0.1 and live\_streaming\_attention > 0.001) \\
\hline
\multirow{2}{*}{$q^{1}_{5}$} & {帮我再查一下市场份额高于均值的吧。} & \multirow{4}{*}{\makecell[c]{Dependent \\ (Ellipsis)}} & \multirow{4}{*}{\makecell[c]{Changing conditions}}\\ ~ & (I want to know the information of the channels whose market share is above average now, can you work out?) \\ \multirow{2}{*}{$y^{1}_{5}$} & {select * from 频道收视 where 市场份额 > ( select avg(市场份额) from 频道收视 )}\\ ~ & (select * from channel\_ratings where market\_share > ( select avg( market\_share ) from channel\_ratings )) \\
\hline
\multicolumn{2}{l}{\textbf{Session 2}}\\
\hline
\multirow{2}{*}{$q^{2}_{1}$} & {将所有基金公司按照注册资金由多到少排一下序，并告诉我它们对应的封闭式与开放式基金之和分别是多少。} & \multirow{4}{*}{\makecell[c]{Independent}} & \multirow{4}{*}{\makecell[c]{--}}\\ ~ & (Sort all funds according to their registered capital from most to least and show me the sum of their closed-end and open-end funds.) \\
\multirow{2}{*}{$y^{2}_{1}$} & {select 名称, 封闭式基金数量 + 开放式基金数量 from 基金公司 order by 注册资本（万） desc}\\ ~ & (select name , sum( num\_closed-end\_funds + num\_open-end\_funds ) from fund\_company order by registered\_capital desc) \\
\hline
\multirow{2}{*}{$q^{2}_{2}$} & {顺便给出它们分别有多少亚洲债券基金吧。} & \multirow{4}{*}{\makecell[c]{Dependent \\ (Both)}} & \multirow{4}{*}{\makecell[c]{Changing SELECT}}\\ ~ & (By the way, how many Asian Bond Fund do they have?) \\ \multirow{2}{*}{$y^{2}_{2}$} & {select 名称, 亚洲债券基金数量, 封闭式基金数量 + 开放式基金数量 from 基金公司 order by 注册资本（万） desc}\\ ~ & (select name , Aisan\_bond\_fund , sum( num\_closed-end\_funds + num\_open-end\_funds ) from fund\_company order by registered\_capital desc
) \\
\hline
\multirow{2}{*}{$q^{2}_{3}$} & {这亚洲债券基金好像没什么参考价值，换成注册资金看看呢？} & \multirow{4}{*}{\makecell[c]{Dependent \\ (Ellipsis)}} & \multirow{4}{*}{\makecell[c]{Changing SELECT}}\\ ~ & (The Asian Bond Fund seems useless, can you replace it with registered captial? ) \\ \multirow{2}{*}{$y^{2}_{3}$} & {select 名称, 注册资本（万）, 封闭式基金数量 + 开放式基金数量 from 基金公司 order by 注册资本（万） desc}\\ ~ & (select name , registered\_captial , sum( num\_closed-end\_funds + num\_open-end\_funds ) from fund\_company order by registered\_capital desc) \\
\hline
\multirow{2}{*}{$q^{2}_{4}$} & {再把上述结果按注册资金从少到多排一下序让我瞅瞅吧。} & \multirow{4}{*}{\makecell[c]{Dependent \\ (Coreference)}} & \multirow{4}{*}{\makecell[c]{Changing display}}\\ ~ & (Put the above result in reverse order and show me that.) \\ \multirow{2}{*}{$y^{2}_{4}$} & {select 名称, 注册资本（万）, 封闭式基金数量 + 开放式基金数量 from 基金公司 order by 注册资本（万） asc}\\ ~ & (select name , registered\_captial , sum( num\_closed-end\_funds + num\_open-end\_funds ) from fund\_company order by registered\_capital asc) \\
\hline
\multicolumn{2}{l}{\textbf{Session 3}}\\
\hline
\multirow{2}{*}{$q^{3}_{1}$} & {电影分很多种类型，麻烦将各类型按对应的电影数从多到少排序。} & \multirow{4}{*}{\makecell[c]{Independent}} & \multirow{4}{*}{\makecell[c]{--}}\\ ~ & (As we all know, there are plenty of movie genres, can you sort them according to the number of movies they have from most to least?) \\
\multirow{2}{*}{$y^{3}_{1}$} & {select 类型 from 电影 group by 类型 order by count(*) desc}\\ ~ & (select genre from movie group by genre order by count(*) desc) \\
\hline
\multirow{2}{*}{$q^{3}_{2}$} & {那是哪种类型一百分钟以上的电影最多？} & \multirow{4}{*}{\makecell[c]{Dependent \\ (Others)}} & \multirow{4}{*}{\makecell[c]{Changing conditions}}\\ ~ & (Which genre has the most movies over 100 minutes?) \\\multirow{2}{*}{$y^{2}_{2}$} & {select 类型 from 电影 where 片长（分钟） > 100 group by 类型 order by count(*) desc limit 1}\\ ~ & (select genre from movie where length > 100 group by genre order by count(*) desc limit 1) \\
\hline
\multirow{2}{*}{$q^{3}_{3}$} & {如果只统计票价超过50元的呢？} & \multirow{6}{*}{\makecell[c]{Dependent \\ (Ellipsis)}} & \multirow{6}{*}{\makecell[c]{Changing conditions}}\\ ~ & (What if we only count movies that cost more than ￥50?) \\ \multirow{4}{*}{$y^{3}_{3}$} & select 电影.类型 from 电影 join 电影上映 on 电影上映.电影id = 电影.词条id where 电影上映.票价（元） > 50 \\ ~ & group by 电影.类型 order by count(*) desc limit 1 \\ ~ & (select movie.genre from movie join movie\_released on movie\_released.id = movie.id where movie\_released.cost > 50 \\ ~ & group by movie.genre order by count(*) desc limit 1)\\
\hline
\multirow{2}{*}{$q^{3}_{4}$} & {又有哪些类型符合条件的电影超过三部？} & \multirow{6}{*}{\makecell[c]{Dependent \\ (Both)}} & \multirow{6}{*}{\makecell[c]{Changing conditions}}\\ ~ & (Which movie genres have more than three movies that meet the criteria?) \\ \multirow{4}{*}{$y^{3}_{4}$} & select 电影.类型 from 电影 join 电影上映 on 电影上映.电影id = 电影.词条id where 电影上映.票价（元） > 50 \\ ~ & group by 电影.类型 having count(*) > 3 \\ ~ & (select movie.genre from movie join movie\_released on movie\_released.id = movie.id where movie\_released.cost > 50 \\ ~ & group by movie.genre having count(*) > 3)\\
\hline
\multicolumn{2}{l}{\textbf{Session 4}}\\
\hline
\multirow{2}{*}{$q^{4}_{1}$} & {社交软件有多少个？} & \multirow{4}{*}{\makecell[c]{Independent}} & \multirow{4}{*}{\makecell[c]{--}}\\~ & (How many social software?) \\
\multirow{2}{*}{$y^{4}_{1}$} & {select count(*) from 社交APP}\\ ~ & (select count(*) from social\_APP) \\
\hline
\multirow{2}{*}{$q^{4}_{2}$} & {占内存30MB以上的呢？} & \multirow{4}{*}{\makecell[c]{Dependent \\ (Ellipsis)}} & \multirow{4}{*}{\makecell[c]{Changing conditions}}\\ ~ & (How about memory usage occupied over 30M?) \\ \multirow{2}{*}{$y^{2}_{2}$} & {select count(*) from 社交APP where 软件大小（M） > 30}\\ ~ & (select count(*) from social\_APP where memory\_usage > 30) \\
\hline
\multirow{2}{*}{$q^{4}_{3}$} & {各公司旗下分别有多少这样的软件？} & \multirow{6}{*}{\makecell[c]{Dependent \\ (Coreference)}} & \multirow{6}{*}{\makecell[c]{Changing conditions}}\\ ~ & (How many software mentioned above does each company have?) \\ \multirow{4}{*}{$y^{4}_{3}$} & select 公司.名称, count(*) from 公司 join 社交APP on 社交APP.母公司id = 公司.词条id where 社交APP.软件大小（M） > 30 \\ ~ & group by 公司.名称 \\ ~ & (select company.name, count(*) from company join social\_APP  on social\_APP.parent\_company\_id=company.id \\ ~ & where social\_APP.memory\_usage > 30 group by company.name) \\
\hline
\multirow{2}{*}{$q^{4}_{4}$} & {分别总共有多少用户注册呢？} & \multirow{6}{*}{\makecell[c]{Dependent \\ (Both)}} & \multirow{6}{*}{\makecell[c]{Changing SELECT}}\\ ~ & (How many users are registered in total respectively?) \\ \multirow{4}{*}{$y^{4}_{4}$} & select 公司.名称, count(*), sum(社交APP.注册用户量（亿）) from 公司 join 社交APP on 社交APP.母公司id = 公司.词条id \\ ~ & where 社交APP.软件大小（M） > 30 group by 公司.名称\\ ~ & (select company.name, count(*), sum(social\_APP.num\_registed\_user) from company join social\_APP  on social\_APP.parent\_company\_id = company.id \\ ~ & where social\_APP.memory\_usage > 30 group by company.name) \\
\hline
%{4-5} 
% &  &  & Previous SQL query $y^i_{j-1}$ & Current SQL query $y^i_{j}$ \\
\end{tabular}
}
\caption{Question sequence examples in Ours.}
\label{tab:more-cases}
\end{table*}

%\section{Example Appendix}
%\label{sec:appendix}

%This is an appendix.
\end{CJK}
\end{document}